# Current Status and Trends in Image Anti-Forensics Research: A Bibliometric Analysis

Yihong Lu[1], Jianyi Liu[1]*, Ru Zhang[1]

[1]School of Cyberspace Security, No. 10 Xitucheng Road, Haidian District Beijing, 100876, Beijing, China

*Corresponding author(s). E-mail(s): liujy@bupt.edu.cn;

Contributing authors: lyhong@bupt.edu.cn; zhang@bupt.edu.cn;

**Abstract**: Image anti-forensics is a critical topic in the field of image privacy and security research. With the increasing ease of manipulating or generating human faces in images, the potential misuse of such forged images is a growing concern. This study aims to comprehensively review the knowledge structure and research hotspots related to image anti-forensics by analyzing publications in the Web of Science Core Collection (WoSCC) database. The bibliometric analysis conducted using VOSViewer software has revealed the research trends, major research institutions, most influential publications, top publishing venues, and most active contributors in this field. This is the first comprehensive bibliometric study summarizing research trends and developments in image anti-forensics. The information highlights recent and primary research directions, serving as a reference for future research in image anti-forensics.

**Keywords**: Deepfake, Image Anti-Forensics, Bibliometric study, Web of Science, Scientometrics

## 1 Introduction

Nowadays, extremely realistic fake images synthesized by Generative Adversarial Network (GAN) [1] have raised widespread public concern and deep apprehension. These concerns primarily revolve around the risk of such fake images being used for illegitimate purposes [2]. For instance, scenes and facial images, meticulously crafted through generative techniques, could be misused to create counterfeit images or videos, leading to the fabrication of false news. Simultaneously, these artificially generated facial images might also be used on social media to impersonate others or to disrupt and deceive biometric recognition security systems. Therefore, researching and developing forensic algorithms capable of effectively detecting these fake images is crucial for ensuring the authenticity and reliability of digital media.

However, some existing forensic algorithms for fake images mainly rely on identifying specific image features or traces, and their accuracy still has limitations [3]. Furthermore, forensic algorithms designed specifically for certain types of image forgery techniques often lack sufficient generalization ability [4]. To overcome these shortcomings, researchers have made numerous attempts. These attempts include the application of deep learning and machine learning technologies to enhance the accuracy and generalizability of forensic algorithms [5], and the development of more complex methods for extracting forgery features to identify more subtle signs of image tampering [6, 7]. Among these, image anti-forensic algorithms deserve particular attention. These algorithms are designed directly to target the potential weaknesses of existing forensic methods, not only revealing the limitations of current technologies but also offering a more proactive and preventative research perspective in the field of digital forensics [8]. Therefore, research on image anti-forensic algorithms is especially important.

Anti-forensics involves the use of certain measures to thwart existing forensic solutions, and its development will in turn promote the advancement of forensic technology. Currently, anti-forensics targeting forged images are mainly divided into two types: reconstruction-based methods and attack methods. Anti-forensics involves attack and reconstruction methods. This section reviews work on attack and image-based reconstruction in this field.

Due to the vulnerability of forensic algorithms based on deep convolutional neural networks to adversarial attacks, most attack methods work by adding imperceptible perturbations to fake images, causing detectors to misclassify. Among the existing deep learning attack methods for fake images, Ding et al. [9] used multiple generators and discriminators to enhance the visual quality of anti-forensic images. Nevertheless, the complex network structure demands substantial computational resources during training. Wang et al. [10] allocated deceptive perturbations in YCbCr color space, but their transferability in other forensic detectors is insufficient.

Most image reconstruction methods based on generative adversarial networks view the regeneration of anti-forensic images as an alignment issue of image content, which intrinsically assumes a similar distribution between fake images and real images. Cui et al. [11] leveraged Sobel filters to learn high-frequency information in images, enhancing anti-forensic performance. Peng et al. [12] proposed a novel generative adversarial network architecture with a unique loss function, enhancing the detail resemblance, like color and lighting, between anti-forensic images and real images. Nguyen et al. [13] utilized an autoencoder to transform fake images into real images and achieved anti-forensics.

Attack anti-forensics often rely on learning aggressive noise interference based on known forensic algorithms. However, the generalization performance of anti-forensics in perturbed images still needs improvement. Image reconstruction methods are achieved by learning the characteristics of real images, but their attack performance is still insufficient. As adversarial perturbations increasingly achieve significant results in attacking deep learning-based forensic algorithms, more and more researchers are beginning to study universal attack-based anti -forensics.

Bibliometrics, as a discipline focused on the analysis of scientific publications, aims to delve into key information of academic papers through quantitative evaluation of scientific research achievements. This method not only examines the frequency and relevance of keywords in research papers but also encompasses a multidimensional analysis of authors, journals, countries, institutions, and references. Thus, it comprehensively evaluates the academic output of a specific research area. Through such multi-faceted analysis, bibliometrics reveals the current state, main hotspots, and development trends of research fields [14, 15]. Moreover, relying on a vast amount of data and empirical research, this method facilitates the further development and refinement of theories through systematic summarization and abstraction. With the massive increase in bibliographic data, this discipline has become crucial for untangling complex research threads and predicting trends. Among the many bibliometric tools, VOSviewer [16] stands out as a key tool for creating scientific knowledge maps based on network data. It combines features from multiple disciplines like visualization technology and computer science, effectively managing and analyzing large-scale bibliographic data. This study employs VOSviewer to analyze the literature in the field of image anti-forensics, aiming to extract valuable insights from the Web of Science-Science Citation Index Expanded (SCI-E) database and to provide directional guidance for future research.

## 2 Data and Methodology

### 2.1 Search Strategy

We utilized the Web of Science Core Collection (WoSCC) to retrieve 1,760 articles related to image anti-forensics. The articles span from 2000 to the present, using a combination of terms and vocabulary associated with image anti-forensics. English search terms included “Anti-Forensic*”, “GAN Generate* image detector*”, “avoid* detect*”, “adversarial image*”, and others. To ensure the comprehensiveness and accuracy of the search, we employed various combinations of these keywords and continuously adjusted and optimized them during the search process. Moreover, to maintain the precision and relevance of the research, we restricted the language of the articles to English. The complete search strategy and the combinations of keywords used are detailed in Table 1.

Table 1 Detailed search strategy for Web of Science Core Collection (WoSCC).

| Search item | Counts |
|---|---|
| TS=(“Anti-Forensic*” OR “Anti-forensic*” OR “anti-Forensic*” OR “anti-forensic*” OR “GAN Generate* image detector*” OR “Generate* image detector*” OR “generate* image detector*” OR “fake image detect*” OR “deepfake image detector*” OR “improve* Fake*” OR “avoid* detect*” OR “dodging detect*” OR “deepfake* evasion” OR “attack* detector*” OR “adversarial image*” OR “impercept* image*”) | 1760 |

2.2 Data Analysis

VOSviewer [16] is adept at effectively displaying complex co-citation networks, encompassing interconnections among paper titles, authors, countries, institutions, and sources of publication. Moreover, VOSviewer also reveals patterns of collaboration among individuals and the temporal trends of different research topics. In the network maps generated by VOSviewer, each node represents a unique entity, such as a country, institution, journal, or author. The size of a node reflects the publication volume of that entity, while the thickness of the lines indicates the strength of the links between entities. Nodes of different colors represent various clusters or time periods. Through this innovative visualization, we delve into the nuances of international and institutional collaborations, explore the symbiotic relationships between journals and their co-cited counterparts, and unravel the intricate web of interactions among authors and their co-cited peers.

Additionally, to gain a more comprehensive understanding of the quality and impact of journals, we referred to the 2023 Journal Citation Reports, paying special attention to the journals' categorization and impact factors. This step is crucial for understanding the dissemination and influence of research findings in the academic community.

**3 Result**

## 3.1 Publication Trend Analysis

In this study, we retrieved a total of 1,760 publications related to image anti-forensics, including 1,730 papers and 30 reviews. These publications exhibit a consistent upward trend, mirroring the rapid advancements in computer vision and deep learning technologies, especially pertinent to image anti-forensics.

Our analysis categorizes these publications into three developmental phases. The initial stage (2000-2005) marks the field's nascent phase, with an average of no more than 10 papers published annually. This was followed by the growth stage (2006-2016), where the field saw a gradual increase in publication volume, averaging up to 100 papers per year. The period witnessed a range between 26 and 93 papers annually. A notable surge in research interest was observed with the advent of Generative Adversarial Network (GAN) technology in 2014. This ushered in the third phase, the explosion stage (2017-2024), characterized by a significant spike in publication volume, averaging about 165 papers per year and peaking at 220 publications in 2021. The proliferation of sophisticated image forgery technologies like Deepfake in 2017 further catalyzed this growth, leading to a 20% increase in 2018 and a 29.7% rise in 2019 compared to the previous year.

As of February 1, 2024, our literature search indicates a potentially lower publication count for 2024, but based on ongoing trends, we anticipate a number comparable to 2023 by year-end (Figure 1). These publications have amassed a total of 23,966 citations, averaging 13.56 citations per paper, which underscores the field's academic impact. Notably, despite the initial low publication volume from 1997 to 2005, there has been a steady increase in both publications and citations, particularly from 2019 to 2023. The year 2021 stands out, with publications cited over 4,000 times, indicative of the heightened interest in image anti-forensics. The field's robust academic activity is further evidenced by an H-index of 71, reflecting the widespread influence of its research findings.

In conclusion, image anti-forensics represents a dynamic and ever-expanding research domain. With ongoing technological advancements and heightened societal concerns over data authenticity, the field's relevance and importance are poised to escalate further.

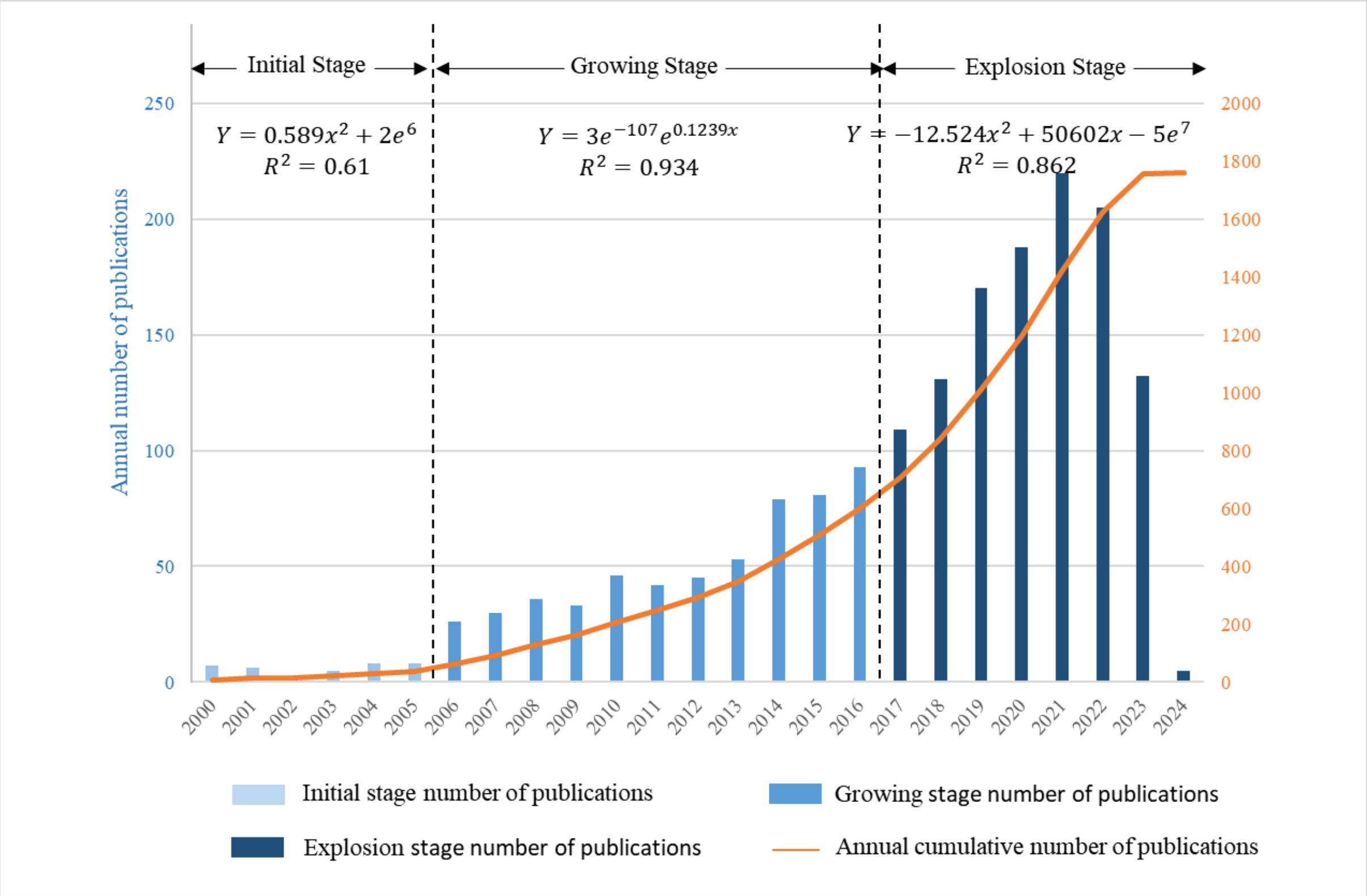

Figure 1 Time Series Analysis: Publication and Citation Dynamics of Image Anti-Forensics Research (2000-2024)

3.2 Cooperation network analysis

3.2.1 Country Analysis

The selected publications originate from 97 countries and 3,412 institutions. The top ten countries in terms of publication volume showcase the diversity and breadth of international research. The United States leads in this field, having published 544 papers, approximately 30.77% of the total, closely followed by China with 393 papers, making up 22.22%. The combined literary output of these two countries accounts for nearly half of the total publications in the field, highlighting their significant global influence in image anti-forensics research (Figure 2).

Further analyzing international collaborations, we created a cooperation network map including countries with a publication count of five or more (Figure 3). This chart highlights the close collaborative relationships between China, the United States, and Germany, but also reflects an uneven distribution of cooperation. The transnational cooperation network still has room for expansion and deepening, particularly in emerging markets and developing countries with high growth potential. To foster comprehensive development in the field of image anti-forensics, broader international cooperation should be encouraged, thus propelling the research in image anti-forensics into a new developmental phase.

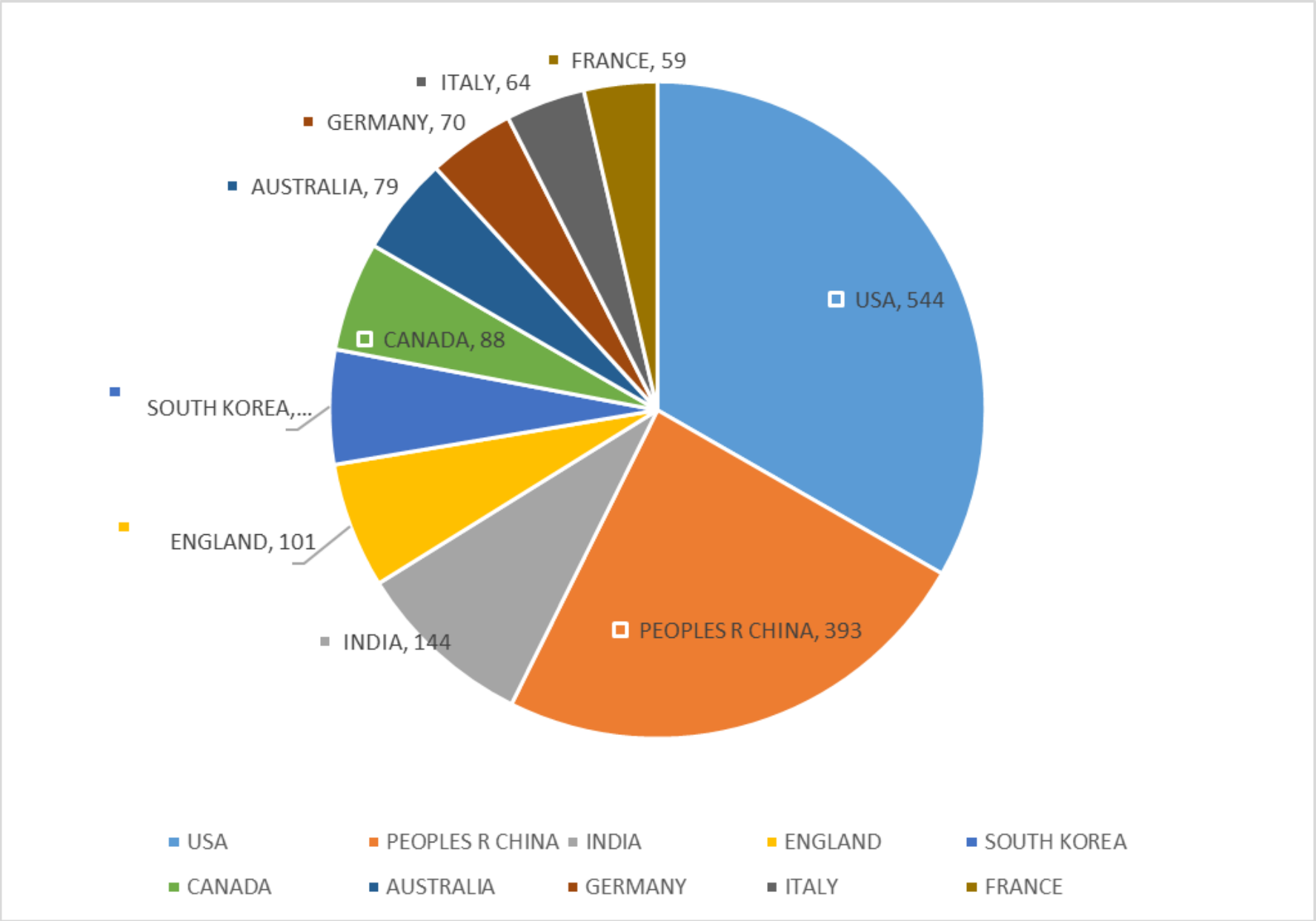

Figure 2 International Distribution of Published Literature in Image Anti-Forensics Research

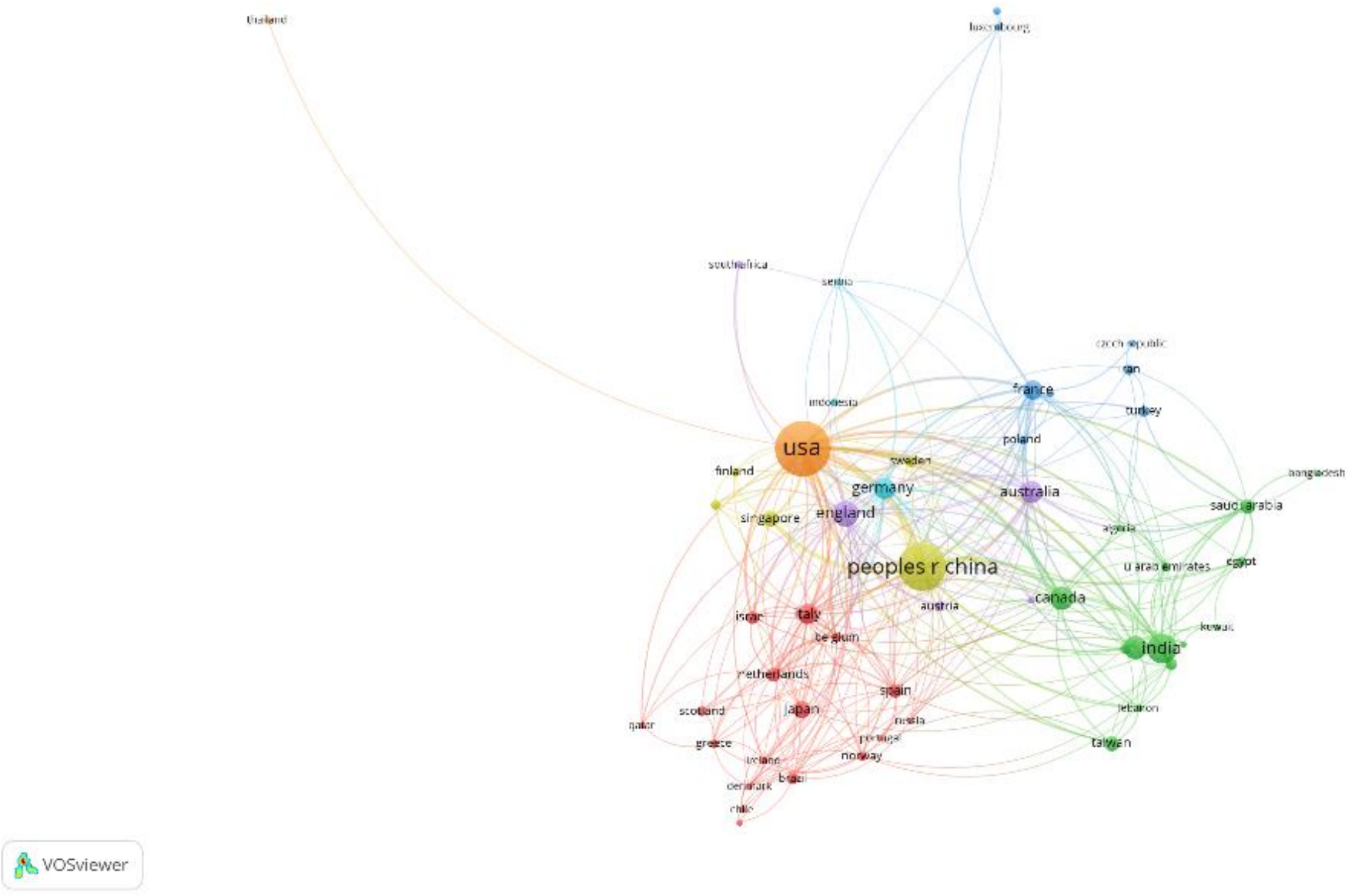

Figure 3 Global Collaboration Network Map for Image Forensics Research.

3.2.2 Institution cooperation network analysis

Looking at the distribution of literature sources, the top ten research institutions that produce the most output span across five different countries. Particularly in China, the Chinese Academy of Sciences leads with 51 publications, demonstrating the vibrancy and influence of its research. Sun Yat-sen University follows closely with 35 publications, while other well-known universities like Tsinghua University also exhibit significant research output (Table 2). Table 3 further reveals the contributions of institutions within China, where institutions such as National University of Defense Technology, Zhejiang University, Shenzhen University, and Beihang University also display strong research capabilities and collaboration potential.

Furthermore, visualizing the cooperation relationships among different countries globally (Figure 4(a)) and among different institutions within China (Figure 4(b)), as shown in the figures, where nodes represent institution names and colors represent time, with yellow indicating recent years' cooperation. From an international perspective, cooperation is particularly close between Chinese Academy of Sciences, Nanyang Technologocal University, Sun Yat-sen University, and Zhejiang University, forming the core of collaboration. In contrast, the University of California system, despite publishing a large volume of research, has relatively fewer cooperation connections in the network. On a domestic scale, a closely-knit cooperation network has formed among institutions such as the Chinese Academy of Sciences, Sun Yat-sen University, National University of Defense Technology, Zhejiang University, Shenzhen University, Tsinghua University, and Beihang University. Although cooperation among domestic institutions is active, the potential for international cooperation has not been fully tapped. In particular, compared to leading international institutions, it can be observed that certain Chinese institutions, while active in research, have insufficient connectivity in the global cooperation network. Therefore, future strategies should include strengthening cooperation and exchange with international research leaders, promoting knowledge complementarity and sharing, and enhancing China's influence and competitiveness in the global field of image anti-forensics research.

Table 2: Ranking of the Top Ten Global Institutions by Publication Volume

| Rank | Country | Publications | Counts |
|---|---|---|---|
| 1 | China | Chinese Academy of Sciences | 51 |
| 2 | USA | University of California System | 39 |
| 3 | China | Sun Yat-sen University | 35 |
| 4 | India | India Institute of Technology System | 33 |
| 5 | France | Centre National De La Recherche Scientifique | 32 |
| 6 | USA | University System of Maryland | 27 |
| 7 | Singapore | Nanyang Technological University | 24 |
| 8 | Singapore | Nanyang Technological University National Institute of Education NIE Singapore | 24 |
| 9 | USA | UNIVERSITY OF TEXAS SYSTEM | 24 |
| 10 | USA | UNIVERSITY SYSTEM OF GEORGIA | 22 |

Table 3: Ranking of the Top Ten Chinese Institutions by Publication Volume

| Rank | Institutions | Counts |
|---|---|---|
| 1 | Chinese Academy of Sciences | 37 |
| 2 | Sun Yat-sen University | 34 |
| 3 | Tsinghua University | 15 |
| 4 | Zhejiang University | 15 |
| 5 | National University of Defense Technology | 15 |
| 6 | Shanghai Jiaotong University | 15 |
| 7 | Shenzhen University | 15 |
| 8 | Beijing Jiaotong University | 13 |
| 9 | Beihang University | 12 |
| 10 | Beijing Institute of Technology | 12 |

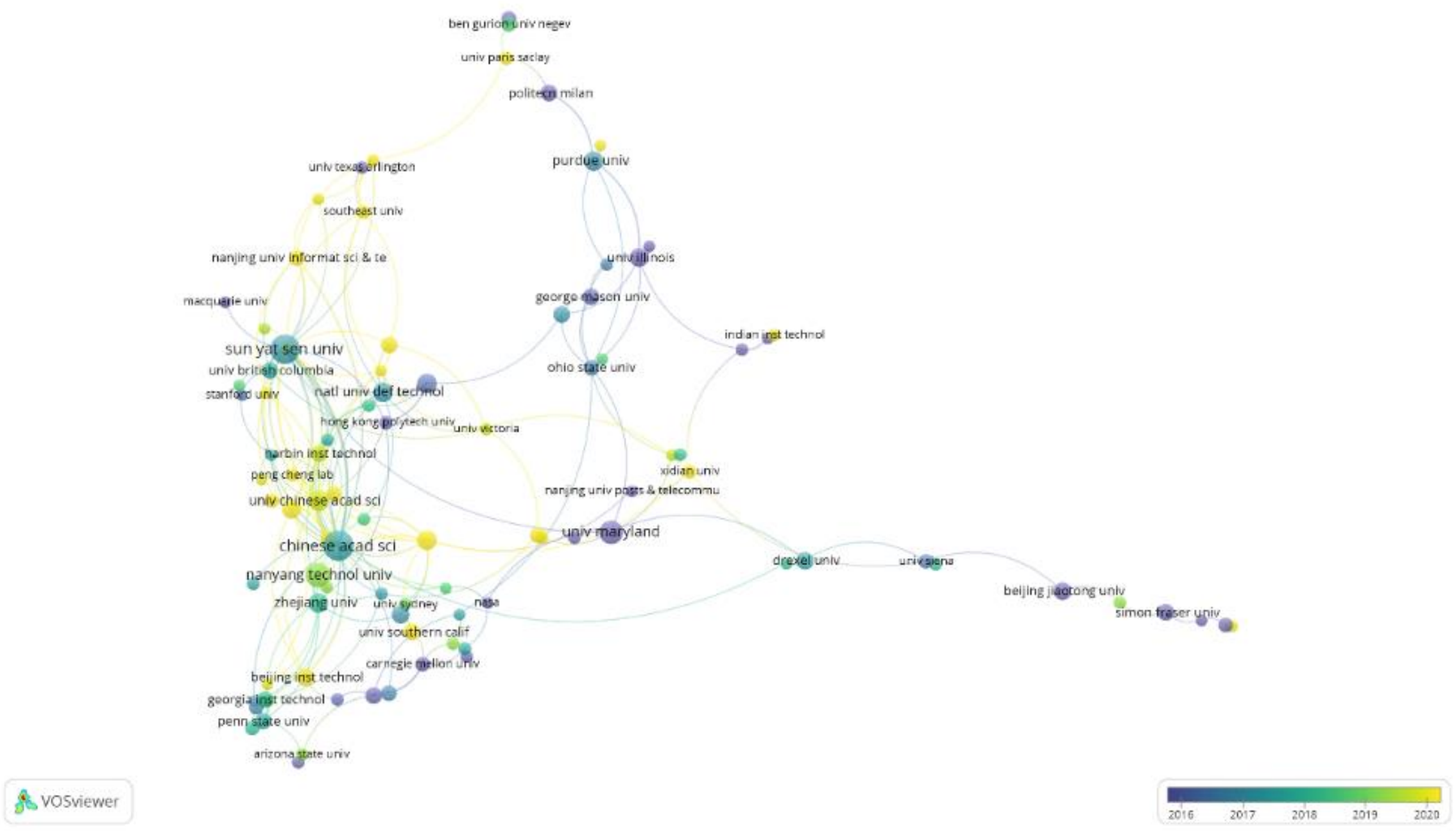

Figure 4(a): International Collaboration Network Map for Image Anti-Forensics Research

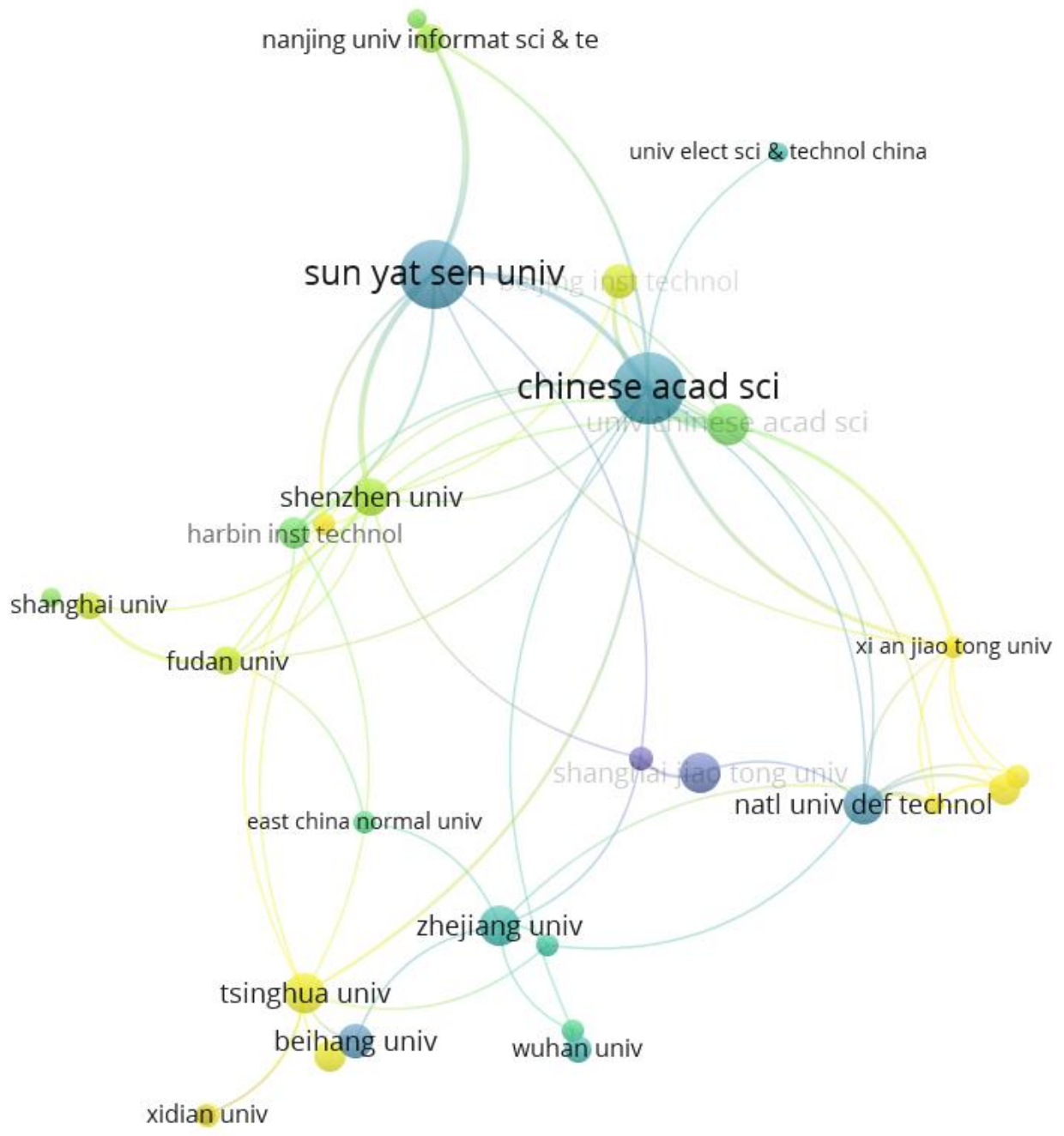

Figure 4(b): Internal Collaboration Network Map for Image Anti-Forensics Research in China

3.2.3 Authors and Co-Cited Authors

In the field of image anti-forensics, a total of 1,968 authors have contributed their research efforts. Firstly, an analysis of prolific authors reveals that among the top 20 authors, the lowest number of published papers is 7 (Table 4). Among these authors, two scholars from Sun Yat-sen University, Kang Xiangui and Zeng Hui, stand out. Their team recently introduced a novel loss function spanning three domains in the field of median filtering image anti-forensics, explore a solution through adversarial learning in an infinite set of possible solutions [17].

Another notable scholar is Matthew Stamm from Drexel University, who has made significant contributions to research in image and video tampering anti-forensics. Stamm, Matthew, in his groundbreaking 2011 paper [8]，first introduced the concept of anti-forensics. This paper, addressing traditional multimedia anti-forensics methods, proposed a series of anti-forensic algorithms based on game theory and core anti-forensic feature statistics. Subsequently, in his 2013 review paper [18], he summarized the theoretical and methodological developments in the emerging field of information anti-forensics over the past decade, highlighted the latest technologies, key applications, and provided insights into future developments. This paper has been cited 229 times to date.

Table 4 Top 10 Authors on Research of Image Anti-Forensics.

| Rank | Author | Counts | Rank | Author | Counts |
|---|---|---|---|---|---|
| 1 | Kang, Xiangui | 20 | 6 | Zhao, Yao | 10 |
| 2 | Stamm, Matthew | 18 | 7 | Ni,Rongrong | 10 |
| 3 | Zeng, Hui | 13 | 8 | Huang, Jiwu | 9 |
| 4 | Liu, K.J. Ray | 13 | 9 | Cao, Gang | 7 |
| 5 | Lee, Sangjin | 11 | 10 | Leprevost, ranck | 7 |

Furthermore, a collaboration network was constructed for 79 authors who have published more than 10 papers in the field of image anti-forensics (Figure 5). It was observed that authors such as Kang Xiangui, Ni Rongrong, Zhang Xinpeng, Stamm Matthew C., and Huang Jiwu formed the most prominent nodes in the network, highlighting their significant research contributions in this field and their close associations with one another. Additionally, Figure 5 also reveals some closely-knit collaborative teams, such as the collaboration group of Kim Jaweon, P.R. Kumar, and Tong Huang, as well as the research team of Kang Xiangui, Peng Manjie, Sun Wei, and Liu Li. Furthermore, Zhang Xinpeng, Lyu Siwei, and Feng Guorio also constitute a closely collaborating group. However, despite the presence of these tightly-knit collaboration groups, the collaboration network among authors still appears relatively closed, with infrequent interactions with the broader academic community. This phenomenon may limit the exchange of new ideas and academic innovation. Therefore, to further advance the field, it is recommended to promote a more open collaboration model, expand interactions among authors, and establish a broader and more diverse collaboration network in the field of image anti-forensics research.

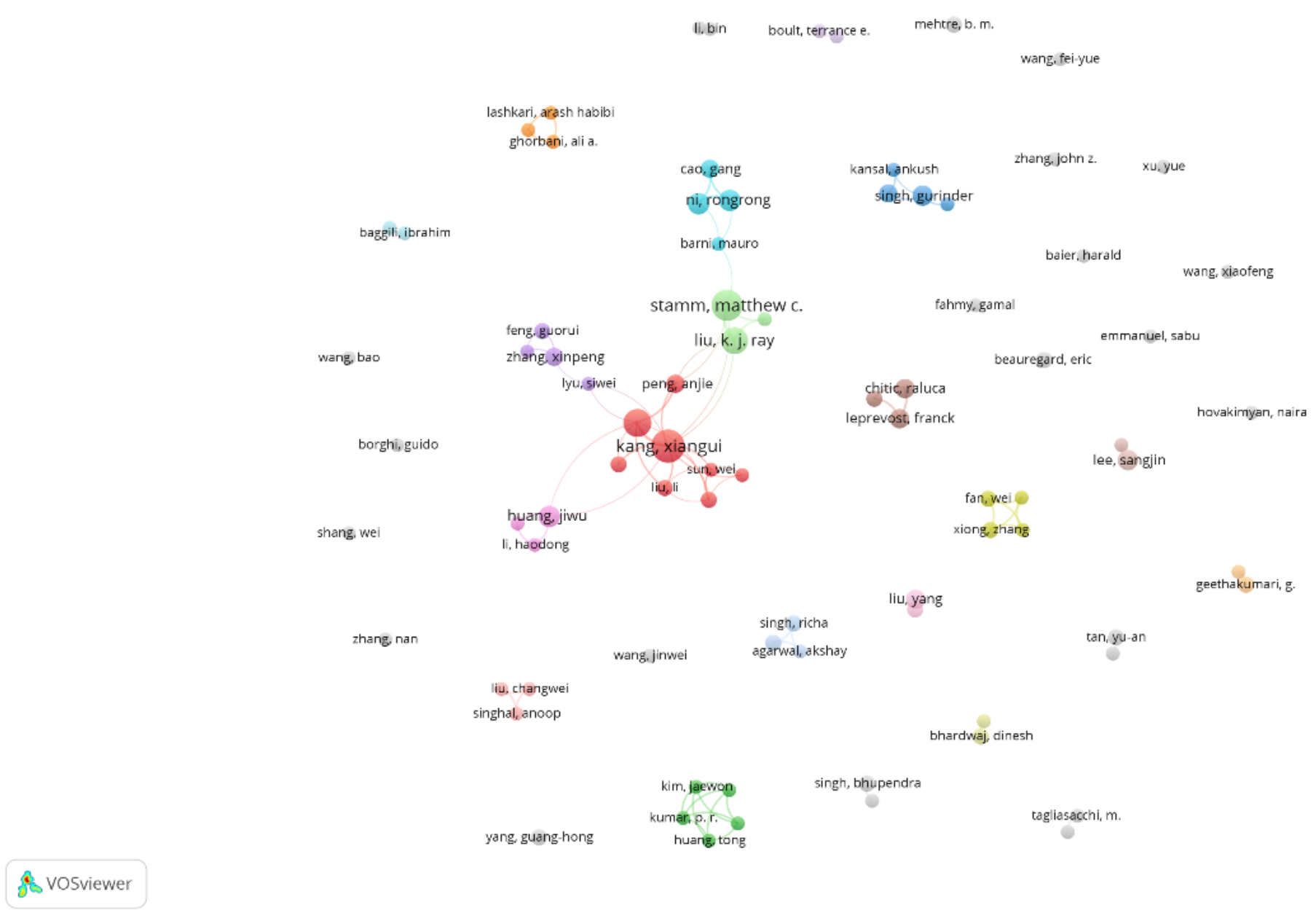

Figure 5 Author Collaboration Network Map

3.3 Knowledge base and research fronts

3.3.1 Category distribution of research

Figure 6 illustrates the distribution of image anti-forensics-related literature across various disciplinary fields, revealing the interdisciplinary nature of this research domain. The most prominent category is 'Engineering Electrical Electronic,' with a total of 620 papers, reflecting the dominant role of electrical engineering in the development of image anti-forensics technologies. Following closely is 'Computer Science Information Systems,' with 474 papers, emphasizing the importance of information systems in data analysis and algorithm applications. The 'Computer Science Theory Methods' category comprises 436 papers, indicating that theoretical research provides the foundational support for image anti-forensics technologies. Additionally, 'Computer Science Artificial Intelligence' has 332 papers, signifying the increasing breadth of artificial intelligence applications in image anti-forensics, especially in machine learning and deep learning technologies. 'Telecommunications' also has 245 papers, highlighting the significance of anti-forensic technologies in image transmission processes. 'Computer Science Software Engineering' with 181 papers reflects the use of software tools in developing anti-forensic technologies. Other categories, including interdisciplinary applications in computer science, hardware architecture, automation control systems, and imaging science and photography technology, have varying numbers of publications. These classifications demonstrate the permeation and application of image anti-forensics technologies across multiple disciplinary fields, from fundamental theoretical research to practical applications.

This diverse disciplinary distribution not only showcases the extensive scope of image anti-forensics research but also signifies a robust, thriving research ecosystem. It highlights the essential role of interdisciplinary collaboration and knowledge exchange in driving the field's growth, paving the way for future research expansion.

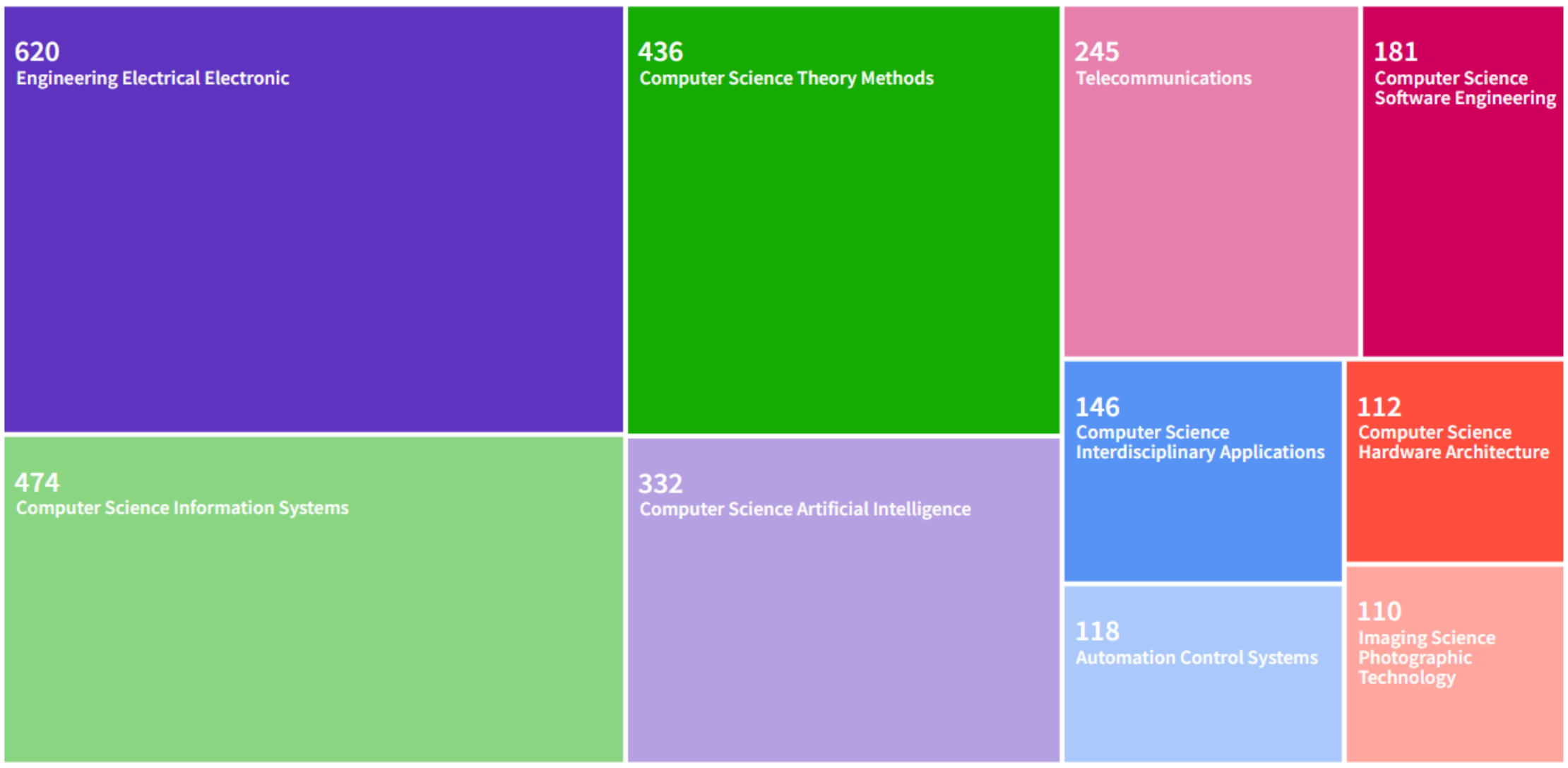

Figure 6: Category Distribution of Image Anti-Forensics Research

3.3.2 Journals and Cited Journals

In the field of image anti-forensics research, relevant literature is widely distributed across 1,353 different journals. The highest number of publications is in 'lecture Notes in Computer Science' (3.45%), demonstrating its central position in academic discourse. Next is 'IEEE Access' (2.20%), followed by 'multimedia Tools and Applications' and 'Proceedings of SPIE' both at 1.92% each. 'IEEE Transactions on Information Forensics And Security' ranks fifth with a publication rate of 1.52%. Among the top 15 journals with the highest number of publications, according to the 2023 Journal Citation Reports, the journals with the highest impact factors are ' IEEE Transactions on Information Forensics and Security ' (IF = 6.8) and 'Neurocomputing' (IF = 6.0). Additionally, important academic conferences such as 'IEEE International Joint Conference on Neural Networks (IJCNN)' (CCF-A), 'communications in Computer and Information Science' (CCF-A), and 'IEEE Conference on Computer Vision and Pattern Recognition' (CCF-A) have been rated as CCF-A level, highlighting their authority in the academic community (Table 5).

Furthermore, using the VOSviewer tool, citation relationships among 175 journals with a publication volume of 5 or more papers were visualized (Figure 7). Notably, there is significant citation interaction between 'CVPR,' ' IEEE Transactions on Information Forensics and Security,' and 'IEEE Access.' This not only confirms their leading positions in image anti-forensics research but also reflects the close connections among them in academic dissemination and knowledge flow, with positive citation relationships among these journals.

Table 5 Top 15 Journals on Research of Image Anti-Forensics.

| Rank | Journal | Count | IF | Q |
|---|---|---|---|---|
| 1 | Lecture Notes in Computer Science | 61 | 0.302 | 4 |
| 2 | IEEE Access | 39 | 4.1 | 2 |
| 3 | Multimedia Tools and Applications | 34 | 3.6 | 3 |
| 4 | Proceedings of SPIE | 34 | - | EI |
| 5 | IEEE Transactions on Information Forensics and Security | 27 | 6.8 | 1 |
| 6 | IEEE International Conference on Image Processing (ICIP) | 26 | - | CCF-C |
| 7 | Digital Investigation | 25 | 2.87 | 3 |
| 8 | Computers Security | 20 | 5.7 | 2 |
| 9 | International Conference on Acoustics Speech and Signal Processing (ICASSP) | 20 | - | CCF-B |
| 10 | Forensic Science International Digital Investigation | 19 | 2 | 3 |
| 11 | IEEE Conference on Computer Vision And Pattern Recognition (CVPR) | 17 | - | CCF-A |
| 12 | Applied Sciences Basel | 15 | 2.9 | 3 |
| 13 | Communications in Computer and Information Science | 13 | - | CCF-A |
| 14 | Neurocomputing | 12 | 6 | 1 |
| 15 | IEEE International Joint Conference in Neural Networks (IJCNN) | 11 | - | CCF-A |

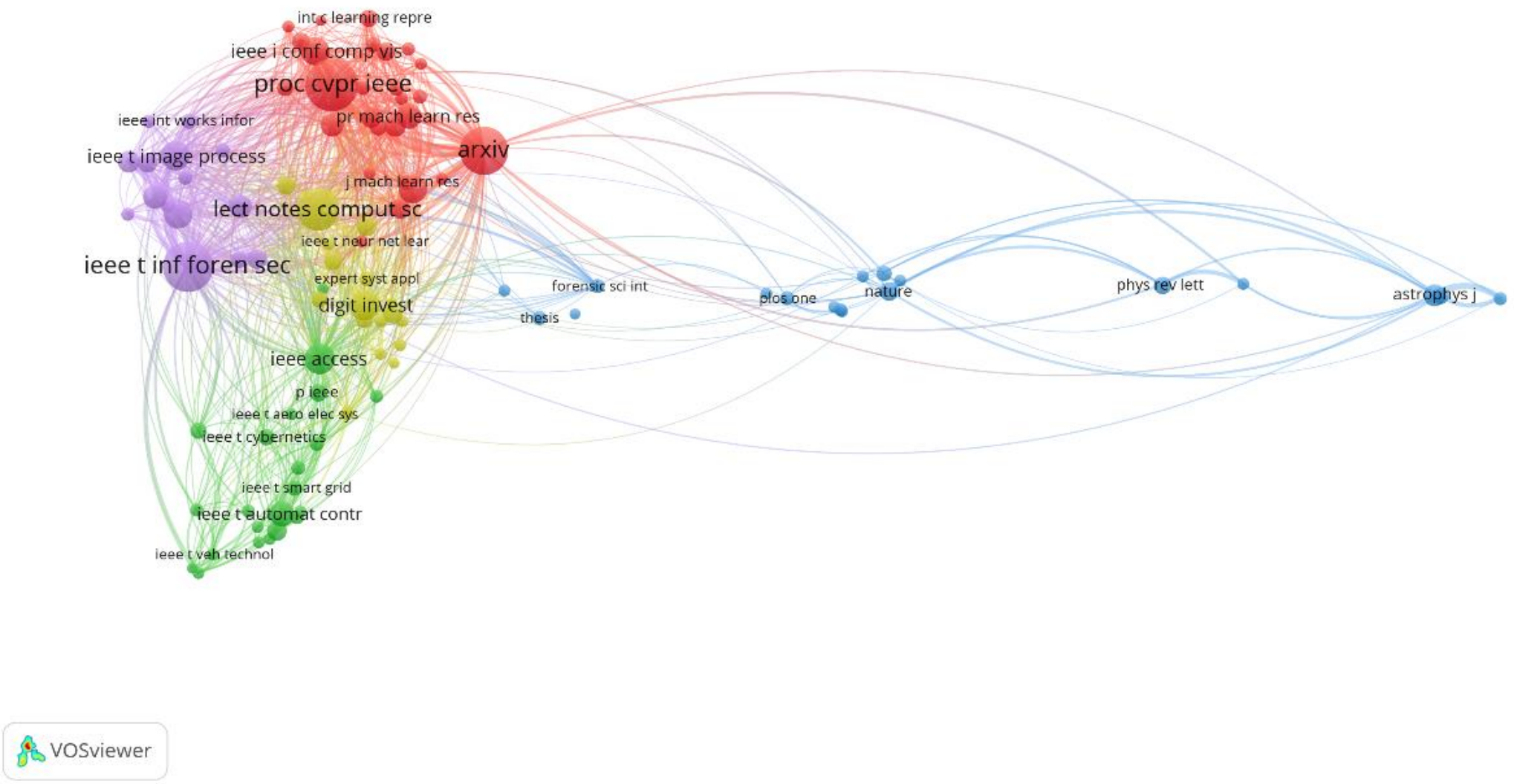

Figure 7 The visualization of Cited Journals on Research of Acupoint Sensitization and Acupoint Specificity

3.3 Hotspots and Frontiers

3.3.1 Keyword co-occurrence clustering analysis

In this study, we conducted co-occurrence analysis of keywords in image anti-forensics-related literature, forming a scientific knowledge map of high-frequency keywords in the field of image anti-forensics. Keywords not only encapsulate the core content of the literature but also indicate the domain of research expertise. In image anti-forensics research, the top 20 high-frequency keywords appeared more than 100 times, collectively forming the primary research directions in this field (Table 6). We further filtered keywords that appeared more than 30 times and performed trend-topic clustering analysis. The color variation within the clusters reflects the temporal relationship of keywords, with closer proximity to yellow indicating more current research frontiers.

Figure 8 illustrates four clusters, signifying three distinct research avenues. The purple cluster, encompassing terms like "anti-forensic," "network forensics," and "game theory," echoes traditional anti-forensics research themes. The green cluster, highlighting "transfer learning," "GAN," and "machine learning," signals the integration of machine learning and GANs in the field. The most recent trends, indicated by the yellow-adjacent cluster, feature "adversarial examples," "defense," and "adversarial learning," underscoring the growing focus on image processing and adversarial learning in contemporary research.

These clusters not only showcase the diversity within image anti-forensics research but also trace the field's evolution. The shift towards advanced data processing and deep learning algorithms, particularly in response to adversarial attack and defense techniques, marks a significant transition from traditional methods to more intricate approaches in recent studies.

Table 6 Top 20 Keywords on Research of Image Anti-Forensics

| Rank | Keywords | Counts | Rank | Keywords | Counts |
|---|---|---|---|---|---|
| 1 | adversarial example | 103 | 11 | time | 300 |
| 2 | robustness | 147 | 12 | technique | 508 |
| 3 | anti-forensic | 146 | 13 | security | 158 |
| 4 | image | 526 | 14 | system | 559 |
| 5 | dataset | 245 | 15 | model | 531 |
| 6 | detection | 934 | 16 | experiment | 291 |
| 7 | attacker | 207 | 17 | evidence | 154 |
| 8 | analysis | 375 | 18 | state | 322 |
| 9 | network | 501 | 19 | study | 338 |
| 10 | effect | 156 | 20 | research | 213 |

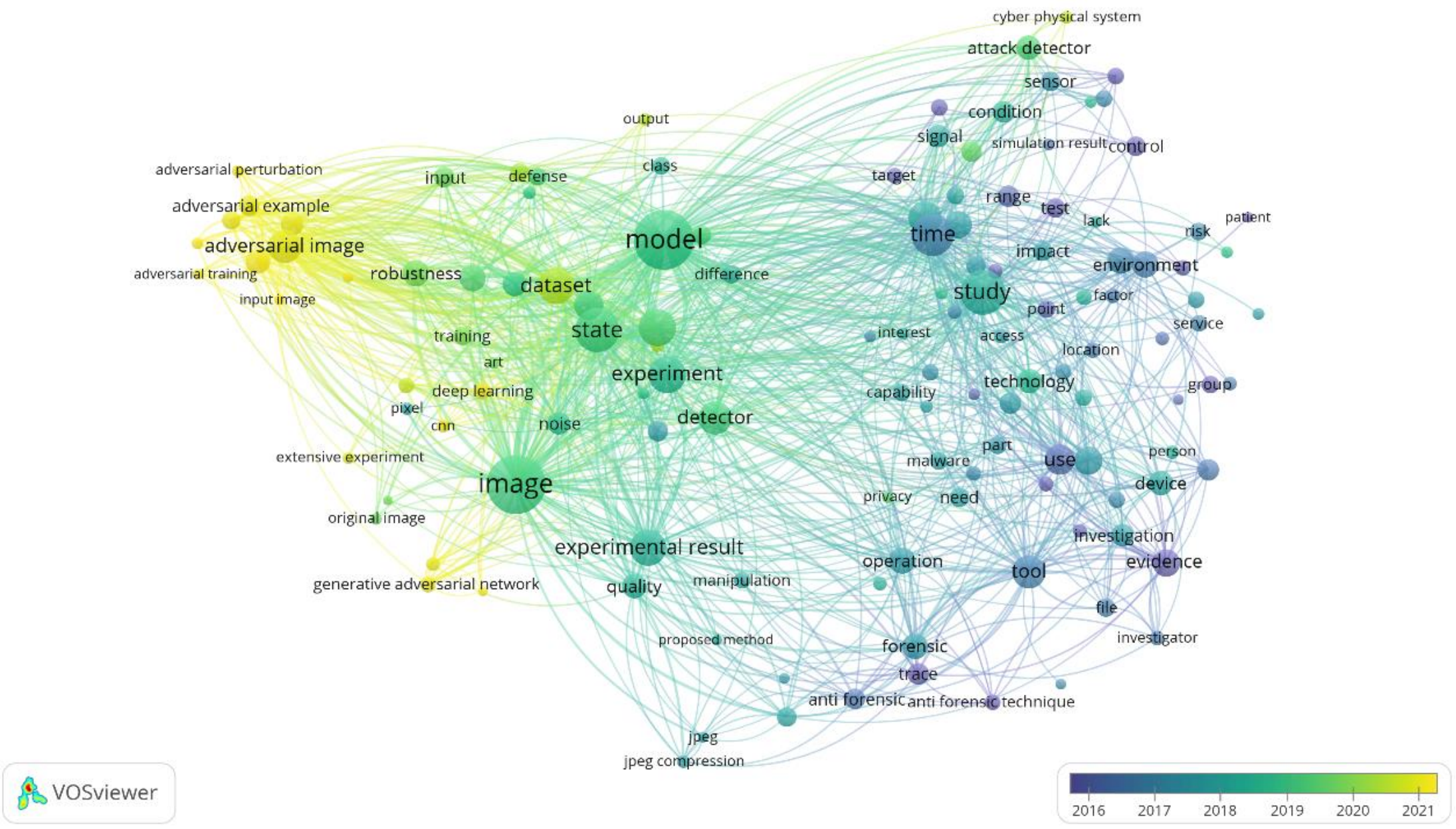

Figure 8 Keyword Cluster Analysis on Research of Image Anti-Forensics

3.3.2 Keyword burst analysis

The keyword burst analysis in image anti-forensics research unveils the evolving and emerging themes within this domain. As depicted in the Figure 9, 'Computer Vision & Graphics' and 'Security Systems' emerge as the most active research areas, constituting 17.421% and 17.138% of all related literature, respectively. These fields underscore their prominence in image anti-forensics. Following closely is 'Security, Encryption & Encoding,' accounting for 15.385%, which emphasizes the growing importance of encryption and encoding in safeguarding image integrity.

Furthermore, 'Artificial Intelligence & Machine Learning' plays a significant role, representing 5.543% of the literature, especially in pattern recognition and automated detection. The fields of 'Automation & Control Systems' and 'Telecommunications' are also notable for their contributions to automated anti-forensics and secure image transmission. While areas like 'Distributed & Real Time Computing' and 'Power Systems & Electric Vehicles' have smaller shares, they highlight their emerging applications in real-time monitoring and mobile device anti-forensics.

This analysis reveals that image anti-forensics research is increasingly intersecting with diverse scientific and engineering disciplines. The identified keyword bursts not only reflect the current research focus but also suggest potential future directions, indicating an ever-expanding interdisciplinary scope.

| Select All ☐ | Field: Citation Topics Meso | Record Count | % of 1,768 |
|---|---|---|---|
| ☐ | 4.17 Computer Vision & Graphics | 308 | 17.421% |
| ☐ | 4.187 Security Systems | 303 | 17.138% |
| ☐ | 4.101 Security, Encryption & Encoding | 272 | 15.385% |
| ☐ | 4.61 Artificial Intelligence & Machine Learning | 98 | 5.543% |
| ☐ | 4.29 Automation & Control Systems | 81 | 4.581% |
| ☐ | 4.13 Telecommunications | 51 | 2.885% |
| ☐ | 4.46 Distributed & Real Time Computing | 44 | 2.489% |
| ☐ | 4.18 Power Systems & Electric Vehicles | 40 | 2.262% |
| ☐ | 4.48 Knowledge Engineering & Representation | 25 | 1.414% |
| ☐ | 1.7 Neuroscanning | 22 | 1.244% |
| ☐ | 4.47 Software Engineering | 21 | 1.188% |
| ☐ | 1.100 Substance Abuse | 20 | 1.131% |
| ☐ | 6.185 Communication | 20 | 1.131% |
| ☐ | 1.21 Psychiatry | 18 | 1.018% |
| ☐ | 4.116 Robotics | 17 | 0.962% |

Figure 9 Keyword Burst Analysis on Research of Image Anti-Forensics

## 4 Discussion

This study, encompassing a comprehensive collection of 1,768 literature records on image anti-forensics from the Web of Science database, spans from 1997 to the present. The early years, from 1997 to 2005, saw a modest annual publication count not exceeding 10 papers, indicating the field's embryonic stage. However, the period between 2006 and 2016 marked a notable upswing in research activity, coinciding with rapid advancements in image processing technology. This era witnessed an average of 51 papers published annually. A pivotal moment occurred in 2017 with the advent of Deepfake technology, catalyzing a surge of interest and research in the field. From 2017 through 2023, the average annual publication count skyrocketed to 165 papers. This significant increase not only reflects the transition of image anti-forensics research into a highly active phase but also highlights its growing prominence as a subject of keen interest in both academic and industrial circles.

In the global landscape of image anti-forensics research, the United States and China emerge as key players, with Chinese institutions notably proactive in international collaborations. Despite a substantial output, some institutions like the University of California system display a degree of independence in the international collaboration network. At the domestic level, Chinese research institutions form strong alliances, yet the extent of their internal cooperation remains limited, suggesting potential areas for broader academic exchange and innovation. In an era of increasing globalization in scientific research, cross-border academic collaboration is indispensable, particularly in a rapidly evolving field like image anti-forensics.

According to the authors, the top three researchers with the highest number of publications are Kang, Xiangui, Stamm, Matthew C., and Zeng, Hui. Kang, Xiangui, and Zeng, Hui are all affiliated with Sun Yat-sen University. Their team has contributed significantly to the field of image steganography and anti-forensics, having published a total of 20 related papers. They began their research in image forensics in 2011 and have presented papers at the Asia-Pacific Signal and Information Processing Association Annual Summit and Conference (APSIPAASC) conference [19]. Additionally, two other prominent researchers with a high publication count are Zhao Y and Ni RR, who are part of a team from Beijing Jiaotong University. They have also published multiple articles in the field of image forensics. Notably, their 2010 paper [20] has been cited in multiple studies related to anti-forensics of median filtering in digital images, with a total of 106 citations.

In the top 15 journals, five of them are conference papers, with three being from top-tier computer vision conferences. This indicates the popularity of conference-type publications in this field, allowing us to quickly track relevant research for the latest results. These conferences are known for publishing cutting-edge advancements and significant breakthroughs in the field of image anti-forensics. Papers presented at CVPR [21], for instance, tend to focus on image anti-forensics from the perspective of adversarial samples. For example, research published in CVPR has revealed that most deep learning-based forensic models are susceptible to simple adversarial attacks. However, existing anti-forensic algorithms that add adversarial noise often suffer from reduced visual quality and are easily detectable. Therefore, instead of adding adversarial noise, researchers have opted to optimally search for adversarial points on the face manifold to generate high-quality anti-forensic fake face images. Furthermore, Byun et al. [22] have approached the issue of adversarial samples from a transferability perspective. To maximize the success rate of transferring adversarial examples, it is crucial to avoid overfitting to the source model. They propose the object-based diverse input (ODI) method, which creates adversarial images on 3D objects, inducing the rendered images to be classified as the target class.

Based on the citation results, it can be observed that the ten most frequently cited papers were published between 2015 and 2018. This period coincides with the rapid growth of computer vision technologies based on deep learning. These papers have been widely referenced in various publications, indicating their significance and impact within the field. Athalye A et al. [23] (cited 292) not only demonstrated the existence of robust 3D adversarial objects but also introduced the first algorithm capable of synthesizing examples that are adversarial across a selected distribution of transformations. The research findings in this paper have made a substantial contribution to the field.

Furthermore, when considering the keyword co-occurrence analysis, it becomes evident that the topic of adversarial attacks began to gain significant attention starting in 2019. This underscores the foundational role and importance of the 2018 paper within the field.

With the advancement of image generation and image forensics technologies, emerging research areas focused on anti-forensic algorithms have garnered increasing attention among researchers. A review of the literature and citation explosion reveals the primary themes of interest in this context: Computer Vision and Graphics, Security System, and Artificial Intelligence and Machine Learning. Concurrently, keyword co-occurrence analysis aids in swiftly identifying the distribution and evolution of research hotspots.

The findings indicate that in the early stages, research in image anti-forensics predominantly centered on concealing tampering and artifacts. Nguyen et al. [13] introduced an approach to enhance the naturalness of fake images from the perspective of detectors. They accomplished this by initially training fake images autoencoders and real image autoencoders and subsequently training a known forensic detectors scheme as a discriminator. Finally, they jointly trained the fake images to real images transformation model, known as H-Net, utilizing two loss functions, resulting in regenerated facial images with well-preserved facial contours. Peng et al. identified certain deficiencies in the attack images generated by existing anti-forensic methods, such as the loss of texture details, variations in lighting, and color distortion. They proposed fake images regeneration GAN (CGR-GAN) [12] based on black-box attacks. In this approach, the generator learns a style mapping function from fake images to real images, approximating the style of fake images to that of real images while retaining the original facial contours. A novel generative adversarial network architecture was introduced to address issues like color deficiency, lack of texture details, and lighting variations in synthetic images. Subsequently, Peng et al. [24] introduced the BDC-GAN structure to achieve bidirectional transformation between natural images and computer-generated images. Zhao et al. [25] achieved robust anti-forensic performance by attacking ensemble detectors with GAN-structured models. Huang et al. [26] mitigated the artifact patterns in GAN-generated images based on dictionary learning. Cui et al. [11] enhanced image contour information by introducing a Sobel filter before one of the discriminator networks in a newly constructed adversarial generation network structure. Their results demonstrated that this method enriches the details of regenerated images, consequently enhancing the efficiency of anti-forensic models.

With a deeper understanding of image generation models, current research in image anti-forensics primarily focuses on adversarial attacks and the generation of adversarial sample images. From the perspective of attacks, the goal is to create carefully crafted perturbations that can render forensic detectors ineffective. Wang et al. [10] applied the Fast Gradient Sign Method (FGSM) and the Momentum Iterative Fast Gradient Sign Method (MI-FGSM) to attack facial images. They observed that a significant portion of the perturbations in the YCbCr color space is concentrated in the Y channel. Consequently, they allocated more perturbations to the Cb and Cr channels to improve the visual quality of anti-forensic images. Goebel et al. [27] employed optimization-based attacks to merge co-occurrence matrices obtained from real images into images generated by GANs. Li et al. [21] conducted an iterative attack known as Projected Gradient Descent (PGD) on the latent vectors and noise of a trained StyleGAN model to directly generate anti-forensic images. Hussain et al. [28] utilized input transformations and PGD attacks to generate adversarial deepfakes with robustness against JPEG compression, achieving promising results.

These efforts from various researchers highlight the importance of developing effective attack strategies to produce high-quality adversarial images and evade forensic detection. As the field of image anti-forensics advances, further exploration and innovation in adversarial attacks are expected to drive the development of more robust anti-forensic techniques.

Our study still has some limitations. Firstly, due to the limitations of the analysis software, the data for this study was extracted solely from one database, Web of Science. This narrow focus may have omitted some relevant articles from other sources, potentially introducing a bias. Secondly, our study only included research published in English. This may have led to the exclusion of publications in languages other than English, potentially introducing a language bias. To address these limitations and provide a more comprehensive overview of the field, future research could consider using multiple databases and including publications in various languages. This would help ensure a more inclusive and diverse representation of the literature on image anti-forensics.

## 5 Conclusion

Image anti-forensics research has significant scholarly value and application prospects. The yearly increase in related literature indicates that the issue of passive forensics and anti-forensics techniques for generated images is gaining widespread attention from researchers both domestically and internationally within the field of multimedia security. Publications from China and the United States are numerous, but there is a need for further strengthening of cooperation and communication between countries and institutions.

On one hand, existing experimental data shows that anti-forensics techniques are significantly effective in hiding or altering traces produced during image generation and editing processes. However, it should be noted that Generative Adversarial Networks (GANs), due to the inherent characteristics of their generator structures, leave specific traces when generating images.

This presents a similar challenge to that of traditional anti-forensic algorithms, which aim to remove target traces while minimizing the introduction of new ones. On the other hand, as forensic technology continues to evolve, updates to forensic models also drive iterations of corresponding anti-forensic models. Therefore, there is an urgent need for in-depth research into the interaction between forensic and anti-forensic technologies to foster a competitive game between them, enabling continuous progress. In particular, constant optimization and updating of these technologies should be pursued through the analysis of existing forensic and anti-forensic strategies.

Future research can focus on the following two aspects: First, in response to the new generation of AI-based image generation and editing technologies, anti-forensic techniques should be designed from an adversarial generation perspective, including the network structure, loss functions, and training strategies, to effectively hide forensic traces across multiple common feature domains. Secondly, as AI technology advances, the automation and optimization of the anonymization process will become key. Research should experiment with facial privacy protection through image anti-forensic technology. The development of data anonymization technology, especially in terms of improving the efficiency and precision of removing or altering sensitive information (such as facial features and geographic location markers) in images, is crucial. This research aims to ensure personal privacy safety while maintaining the practical value of image data for secure use in public research and data sharing.

Reference

[1] Ian J. Goodfellow, Jean Pouget-Abadie, Mehdi Mirza, Bing Xu, David Warde-Farley, Sherjil Ozair, Aaron Courville, and Yoshua Bengio. 2014. Generative adversarial nets. In Proceedings of the 27th International Conference on Neural Information Processing Systems - Volume 2 (NIPS'14). MIT Press, Cambridge, MA, USA, 2672–2680.

[2] El-Shafai, W., Fouda, M.A., El-Rabaie, ES.M. et al. A comprehensive taxonomy on multimedia video forgery detection techniques: challenges and novel trends. Multimed Tools Appl 83, 4241–4307 (2024). https://doi.org/10.1007/s11042-023-15609-1.

[3] F. Marra, D. Gragnaniello, D. Cozzolino, and L. Verdoliva, “Detection of GAN-Generated Fake Images over Social Networks,” in MIPR, 2018. 1, 2, 4, 5

[4] Zhang X, Karaman S, Chang S F. Detecting and simulating artifacts in gan fake images[C]//2019 IEEE International Workshop on Information Forensics and Security (WIFS). IEEE, 2019: 1-6.

[5] Qian Y, Yin G, Sheng L, et al. Thinking in frequency: Face forgery detection by mining frequency-aware clues[C]//European Conference on Computer Vision.Springer, 2020: 86-103.

[6] Afchar D, Nozick V, Yamagishi J, et al. Mesonet: a compact facial video forgery detection network[C]//2018 IEEE International Workshop on Information Forensics and Security (WIFS). IEEE, 2018: 1-7.

[7] Yu N, Davis L S, Fritz M. Attributing fake images to gans: Learning and analyzing gan fingerprints[C]//Proceedings of the IEEE International Conference on Computer Vision. 2019: 7556-7566.

[8] Stamm M C and Liu K J R. 2011. Anti-forensics of digital image compression. IEEE Transactions on Information Forensics and Security, 6(3): 1050-1065 [DOI: 10. 1109 / TIFS. 2011. 2119314]

[9] F. Ding, G. Zhu, Y. Li, X. Zhang, P. K. Atrey and S. Lyu, “Anti-Forensics for Face Swapping Videos via Adversarial Training,” IEEE Transactions on Multimedia, vol. 24, pp. 3429-3441, 2022.

[10] Y. Wang, X. Ding, Y. Yang, L. Ding, R. Ward, and Z. J. Wang, “Perception matters: Exploring imperceptible and transferable anti-forensics for GAN-generated fake face imagery detection,” Pattern Recognit. Lett., vol. 146, pp. 15-22, Jun. 2021.

[11] Cui Q, Meng R, Zhou Z, “An anti-forensic scheme on computer graphic images and natural images using generative adversarial networks,” Mathematical Biosciences and Engineering, 2019, 16(5):4923-4935.

[12] Peng F, Yin L P, Zhang L B and Long M. 2020. CGR-GAN: CG facial image regeneration for anti-forensics based on generative adversarial network. IEEE Transactions on Multimedia, 22 (10): 2511-2525

[13] H. H. Nguyen, N. -D. T. Tieu, H.-Q. Nguyen-Son, J. Yamagishi and I. Echizen, “Transformation on Computer-Generated Facial Image to Avoid Detection by Spoofing Detector,” 2018 IEEE International Conference on Multimedia and Expo (ICME), San Diego, CA, USA, 2018, pp. 1-6.

[14] Wang, B., Xing, D., Zhu, Y., Dong, S., and Zhao, B. (2019). The state of exosomes research: a global visualized analysis. Biomed. Res. Int. 2019, 1–10. doi: 10.1155/2019/1495130

[15] Ke, L., Lu, C., Shen, R., Lu, T., Ma, B., and Hua, Y. (2020). Knowledge mapping of drug-induced liver injury: a scientometric investigation (2010-2019). Front. Pharmacol.11:842. doi: 10.3389/fphar.2020.00842

[16] van Eck, N. I., and Waltman, L. (2010). Software Survey: VOSviewer, a Computer Program for Bibliometric Mapping. Scientometrics 84 (2), 523 – 538. doi:10.1007/s11192-009-0146-3

[17] Wu, Jianyuan, et al. "An adversarial learning framework with cross-domain loss for median filtered image restoration and anti-forensics." Computers & Security 112-(2022):112.

[18] M. C. Stamm, M. Wu and K. J. R. Liu, "Information Forensics: An Overview of the First Decade," in IEEE Access, vol. 1, pp. 167-200, 2013, doi: 10.1109/ACCESS.2013.2260814.

[19] X. Kang, M. C. Stamm, A. Peng and K. J. R. Liu, "Robust median filtering forensics based on the autoregressive model of median filtered residual," Proceedings of the 2012 Asia Pacific Signal and Information Processing Association Annual Summit and Conference, Hollywood, CA, USA, 2012, pp. 1-9.

[20] G. Cao, Y. Zhao, R. Ni, L. Yu and H. Tian, "Forensic detection of median filtering in digital images," 2010 IEEE International Conference on Multimedia and Expo, Singapore, 2010, pp. 89-94, doi: 10.1109/ICME.2010.5583869.

[21] D. Li, W. Wang, H. Fan and J. Dong, "Exploring Adversarial Fake Images on Face Manifold," 2021 IEEE/CVF Conference on Computer Vision and Pattern Recognition (CVPR), Nashville, TN, USA, 2021, pp. 5785-5794, doi: 10.1109/CVPR46437.2021.00573.

[22] J. Byun, S. Cho, M. -J. Kwon, H. -S. Kim and C. Kim, "Improving the Transferability of Targeted Adversarial Examples through Object-Based Diverse Input," 2022 IEEE/CVF Conference on Computer Vision and Pattern Recognition (CVPR), New Orleans, LA, USA, 2022, pp. 15223-15232, doi: 10.1109/CVPR52688.2022.01481.

[23] Athalye A , Engstrom L , Ilyas A ,et al.Synthesizing Robust Adversarial Examples[J]. 2017.DOI:10.48550/arXiv.1707.07397.

[24] F. Peng, L. -P Yin, M. Long. et al. “BDC-GAN: Bidirectional Conversion between Computer-generated and Natural Facial Images for Anti-forensics,” IEEE Transactions on

Circuits and Systems for Video Technology, vol.32, no. 10, pp. 6657-6670, Oct. 2022.

[25] X. Zhao and M. C. Stamm, “Making generated images hard to spot: A transferable attack on synthetic image detectors,” 2021, arXiv:2104.12069.

[26] Huang, M. -Y. Liu, S. Belongie, and J. Kautz, “Multimodal unsupervised image-to-image translation,” in Proc. Eur. Conf. Comput. Vision, 2018, pp. 172-189.

[27] M. Goebel and B. S. Manjunath, “Adversarial attacks on co-occurrence features for GAN detection,” 2020, arXiv:2009.07456.

[28] S. Hussain et al., “Adversarial DeepFakes: Evaluating vulnerability of deepfake detectors to adversarial examples,” in Proc. IEEE/CVF Winter Conf. Appl. Comput. Vis., Jan. 2021, pp. 3348–3357.